\documentclass{article}

\usepackage{conf/PRIMEarxiv}

\usepackage[utf8]{inputenc} 
\usepackage[T1]{fontenc}    
\usepackage{hyperref}       
\usepackage{url}            
\usepackage{booktabs}       
\usepackage{amsfonts}       
\usepackage{nicefrac}       
\usepackage{microtype}      
\usepackage{xcolor}         

\usepackage[utf8]{inputenc} 
\usepackage[T1]{fontenc}    
\usepackage{hyperref}       
\usepackage{url}            
\usepackage{booktabs}       
\usepackage{amsfonts}       
\usepackage{nicefrac}       
\usepackage{microtype}      
\usepackage{xcolor}         
\usepackage[table]{xcolor}
\usepackage{multirow}
\usepackage{graphicx}
\usepackage{xspace}
\usepackage{amsmath}
\usepackage{amsthm}
\usepackage{mathtools}
\usepackage{soul}
\usepackage{algorithm}
\usepackage{algpseudocode}
\usepackage{caption}

\usepackage{color}

\makeatletter
\DeclareRobustCommand\onedot{\futurelet\@let@token\@onedot}
\def\@onedot{\ifx\@let@token.\else.\null\fi\xspace}

\def\ie{\emph{i.e}\onedot} 
 
\def\etc{\emph{etc}\onedot}

\makeatother

\definecolor{mygray}{RGB}{224,224,224}

\newif\ifrevision
\revisiontrue

\ifrevision
  
\else
  
\fi

\title{Intermediate Representations are Strong AI-Generated Image Detectors}

\author{
  Zhenhan Huang \\
  Department of Computer Science \\
  Rensselaer Polytechnic Institute \\
  Troy \\
   \And
  Pin-Yu Chen, Tejaswini Pedapati\\
  IBM research \\
  Yorktown \\
   \And
  Jianxi Gao \\
  Department of Computer Science \\
  Rensselaer Polytechnic Institute \\
  Troy \\
}

\begin{document}

\maketitle

\begin{abstract}
    The rapid advancement in generative AI models has enabled the creation of photorealistic images. At the same time, there are growing concerns about the potential misuse and dangers of generated content, as well as a pressing need for effective AI-generated image detectors. However, current training-based detection techniques are typically computationally costly and can hardly be generalized to unseen data domains, while training-free methods fall short in detection performance. To bridge this gap, we propose a search-based method employing data embedding sensitivity in intermediate layers to detect AI-generated images. Given a set of real and AI-generated images, our method examines the similarity between original image embeddings and perturbed image embeddings, and detects AI-generated images based on the similarity. We examine the proposed method on two comprehensive benchmarks: GenImage and Forensics Small. Our method exhibits improved performance across different datasets compared to both training-free and training-based state-of-the-art methods. On average, our method achieves the largest performance gain on the Forensics Small benchmark by 39.61\% compared to the best training-free method and 5.14\% compared to the best training-based method in AUROC score.
\end{abstract}

\section{Introduction}

The advent of image generative models enables the creation of realistic synthetic images. Fueled by advances in deep learning techniques, generative models such as generative adversarial network (GAN) \cite{goodfellow2020generative,metz2016unrolled,liu2016coupled,mao2017least,yoon2019time,karras2019style}, Variational Autoencoder (VAE) \cite{mescheder2017adversarial,mishra2018generative,pinheiro2021variational,he2022masked}, diffusion model \cite{ho2020denoising,song2020denoising,saharia2022photorealistic,podell2023sdxl,blattmann2023align,peebles2023scalable}, \etc. have demonstrated significant progress in image generation. While some image-generation applications have attracted users to go bananas, generative models pose serious ethical, societal, and security challenges. The misuse and the associated cost of generated images can cause negative impacts such as copyright violation, deepfake, and fake content in publications. Furthermore, training datasets for deep learning models might be corrupted by generated images at scale, leading to unintentional bias or malicious exploits for future models. These critical challenges underscore the need for reliable AI-generated image detection.

\begin{figure}[t]
    \centering
    \includegraphics[width=.9\linewidth]{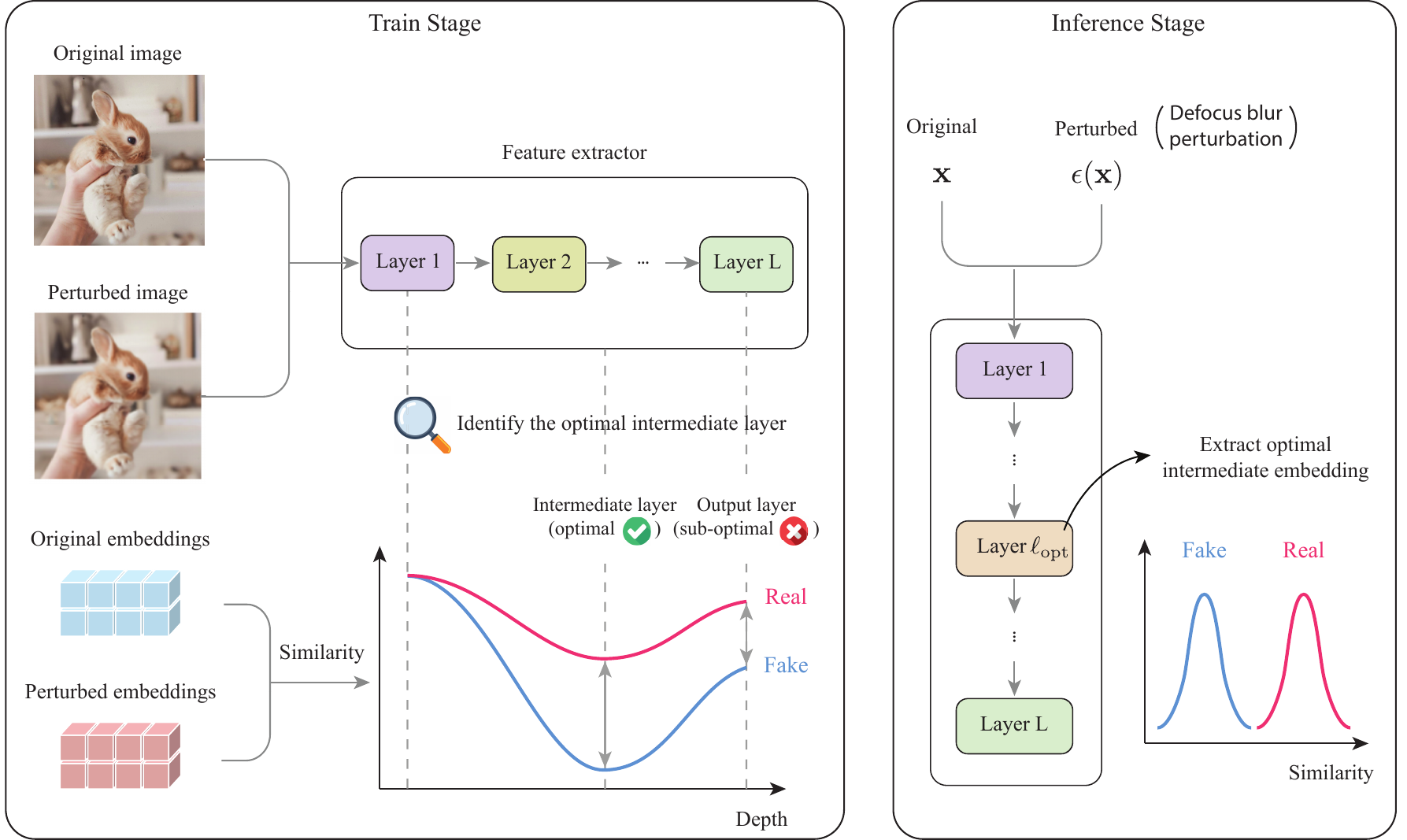}
    \caption{Illustration of the proposed method. The framework consists of two stages. In the train stage, both the original image and the perturbed image are fed to the feature extractor (a pretrained image foundation model). Embeddings across all layers are extracted to obtain intermediate representations. The cosine similarity between the embeddings of the original image and the perturbed image is computed as the metric to make a binary classification on whether an image is AI-generated. In the inference stage, we use the optimal intermediate embedding for the detection, which yields a better detection than the output embedding. }
    \label{fig:illust}
\end{figure}

There are two mainstream approaches to detecting AI-generated images: \textit{training-based} and \textit{training-free} approaches. Current training-based approaches have limited generalization to unseen data domains, while training-free approaches have inferior detection performance.
To bridge the gap, we propose a simple yet effective search-based detector that exploits a pre-trained image foundation model to detect AI-generated images. Figure~\ref{fig:illust} shows the workflow of the proposed method. Following prior arts of training-free detection \cite{he2024rigid,tsai2024understanding} that use a similarity score computed by a pair of test image and its perturbed version for detection, our method examines the robustness of image embedding space to detect AI-generated images. However, different from the prior works that focus on the output embedding, our method utilizes the intermediate embedding space. Our method also differs from training-based detection methods in that we use some training data to \textit{search} for the best intermediate layer for detection, instead of modifying the weights of the underlying pre-trained model for feature extraction.  
Specifically, a pair of original image and perturbed image are fed into a feature extractor. The cosine similarity between the pair of image embeddings is computed across model layers. By employing the embeddings of the optimal intermediate layer, AI-generated images can be detected, assuming the embedding space of the real images is more robust compared to that of AI-generated images. A similarity threshold value is used for the binary classification to determine if an image is AI-generated.

Our method is inspired by the analysis result of the dimensionality of data manifolds: we empirically observe the dimensionality expansion and compression across the depth of deep learning models. The evolution of the geometry of the representation manifold reveals more expressive embeddings of intermediate layers, which is beneficial to detect nuances between real and AI-generated images. Our method exhibits better performance compared to both training-free approaches and training-based approaches on the GenImage benchmark \cite{zhu2023genimage} and the Forensics Small benchmark \cite{park2025community}.

\section{Related Work}

\paragraph{AI-Generated Image Detection} Frequency domain analysis is found to be effective to detect AI-generated images \cite{frank2020leveraging,chandrasegaran2021closer,corvi2023intriguing}. In addition to handcrafted features, learning-based methods are proposed to exploit the strength of neural networks \cite{corvi2023detection,cozzolino2021towards,gragnaniello2021gan,ojha2023towards}. UniDetector \cite{ojha2023towards} uses both nearest neighbor (training-free) and linear probing (training-based) on the image embedding space to detect AI-generated images. NPR \cite{tan2024rethinking} trains a detector that is generalizable to detect images generated by both GANs and diffusion models. The detector relies on neighboring pixel relationships based on the observation that local independence among image pixels exhibits generalized forgery artifacts in generated images. AIDE \cite{yan2024sanity} captures both low-level pixel statistics and high-level global semantics to detect anomalies in AI-generated images such as white noise in the image (low level) and unreasonable image components in the context (high level). SPAI \cite{karageorgiou2025any} uses spectral learning to distinguish AI-generated images based on the spectral reconstruction similarity.

In addition to learning-based methods, training-free methods, not limited to the training dataset, are proposed. AeroBlade \cite{ricker2024aeroblade} assumes that the reconstruction of AI-generated images is easier than that of real images. Hence, the reconstruction error can be used as the metric to detect AI-generated images. RIGID \cite{he2024rigid} assumes that AI-generated images are less robust to perturbations in the embedding space of neural architectures. MINDER \cite{tsai2024understanding} improves the prediction of the RIGID method by introducing contrastive perturbation. Both MINDER and RIGID methods rely on the output (final) layer's embedding of the feature extractor to detect AI-generated images.

\paragraph{Exploiting Intermediate Layers} Intermediate layers are found to be able to enhance the prediction and assist in the analysis of neural architectures. They are used to predict generalization gaps \cite{jiang2018predicting}, elucidate training dynamics through linear classifier probes \cite{alain2016understanding}, improve transfer learning \cite{evci2022head2toe}, enhance the adversarial example transferability \cite{huang2019enhancing}, and ameliorate the performance of fine-tuned models \cite{lee2022surgical}. A fundamental geometric property of the data representation in over-parameterized neural networks is the \textit{intrinsic dimension}, \ie the minimal number of coordinates necessary to describe data points without significant information loss. It is found that the intrinsic dimension increases in earlier layers (expansion) and decreases in later layers (compression) \cite{ansuini2019intrinsic,recanatesi2019dimensionality}. 

\section{Intermediate Representations as AI-Generated Image Detectors}
The overall flow of this section is as follows: First, we formally define the task formulation of our search-based detection framework. Then, we present our proposed method. Next, we provide motivating examples to articulate the importance of selecting the optimal intermediate layer to obtain discriminative features for detection. Finally, we explain why intermediate representations are powerful features for AI-generated image detection through the lens of intrinsic dimension analysis.

\paragraph{Task Formulation} Given a set of labeled images $\mathcal{D} = \{(\mathbf{x}_i, y_i)\}_{i=1}^{n}$ with $\mathbf{x}_i \in \mathcal{X}$ denoting an image and $y_i \in \{0, 1\}$ denoting its label. $y_i = 1$ indicates AI-generated image while $y_i = 0$ indicates real image. Using a pretrained image feature extractor $\mathcal{F}(\cdot)$, the goal is to assign a predicted label $\hat{y}$ for a test image $\mathbf{x}$.

The aim of this paper is to explore the potential of intermediate representations for search-based AI-generated image detection. We use a feature extractor for inference purpose only. Both the model weight and architecture stay the same during the entire detection process.

\subsection{Proposed Method}

Figure \ref{fig:illust} shows the illustration of the proposed method. We feed a pair of the original image $\mathbf{x} \in \mathcal{X}$ and the perturbed image $\epsilon(\mathbf{x}) \in \mathcal{X}$ to the model $\mathcal{F} = f_{L} \circ \ldots \circ f_1$, where $f_{\ell}$ denotes the $\ell$-th layer of $\mathcal{F}$. We use the training dataset to determine the optimal layer $\ell_{\rm opt}$.

Both $\mathbf{x}$ and $\epsilon(\mathbf{x})$ constitute a pair to compute the cosine similarity that characterizes the drift in the embedding space caused by a perturbation. We extract embeddings in the $\ell$-th intermediate layer $\mathcal{F}_{\rm sub} = f_{\ell_{\rm opt}} \circ \ldots f_1$, $1 \le \ell_{\rm opt} \le L$, and compute the cosine similarity between the embeddings of the original image and the perturbed image. Let $\text{emb}(\cdot)$ denote the function to extract the class embedding $\mathbf{e}^{(\ell_{\rm opt})} \in \mathbb{R}^d$ as the intermediate representation for each layer. For example, in DINOv2 \cite{oquab2023dinov2} and CLIP \cite{radford2021learning}, $\text{emb}(\cdot)$ extracts \texttt{[CLASS]} token embedding. 
  The cosine similarity for the optimal layer $\ell$ is defined as
\begin{equation}\label{eq:metric}
\begin{split}
    & \mathbf{e}^{(\ell_{\rm opt})}_{\rm org} = \text{emb}(f_{\ell_{\rm opt}} \circ \ldots\circ f_1(\mathbf{x})) ,\; \mathbf{e}^{(\ell_{\rm opt})}_{\rm perturb} = \text{emb}(f_{\ell_{\rm opt}} \circ \ldots \circ f_1(\epsilon(\mathbf{x})), \\
    & S(\mathbf{x}, \epsilon(\mathbf{x}), \ell_{\rm opt}) = \text{sim}(\mathbf{e}^{(\ell_{\rm opt})}_{\rm org}, \mathbf{e}^{(\ell_{\rm opt})}_{\rm perturb}), \text{ where } \text{sim}(\mathbf{v}_1, \mathbf{v}_2) = \frac{\langle \mathbf{v}_1, \mathbf{v}_2 \rangle}{\lVert \mathbf{v}_1 \rVert \lVert \mathbf{v}_2 \rVert}.\\
\end{split}
\end{equation}
Here $\langle \cdot , \cdot \rangle$ denotes the inner product of two vectors, and $\|\cdot \|$ is the Euclidean norm. $d$ is the hidden dimension defined in the feature extractor.

A threshold value $\tau$ is used to determine whether an image is AI-generated. Our search space assumes that real image embedding space is more robust than that of AI-generated image embedding space. Upon determining the optimal layer $\ell_{\rm opt}$, we can conclude that if $S(\mathbf{x}, \epsilon(\mathbf{x}), \ell_{\rm opt}) < \tau$, the input image is AI-generated. Hence, the prediction logit can be determined by:
\begin{equation}\label{eq:infer}
    y = \mathbb{I}\{S(\mathbf{x}, \epsilon(\mathbf{x}), \ell_{\rm opt}) < \tau\},
\end{equation}
where $\mathbb{I}\{\cdot\}$ is the indicator function and $\mathbb{I}\{\mathcal{A}\} = 1$ if and only if an event $\mathcal{A}$ happens.

Algorithm \ref{alg:method} depicts the pipeline for detecting AI-generated images. There are two stages: in stage I, we determine the optimal layer that maximizes the AUROC score using the training dataset $\mathcal{D}_{\rm tr}$. In the stage II, we make a prediction using the optimal layer representation.

\begin{algorithm}[tbh]
    \caption{Using intermediate representations to detect AI-generated images}
    \label{alg:method}
    \begin{algorithmic}
    \State {\bfseries Input:} Randomly sampled training dataset $\mathcal{D}_{\rm tr} = \{(\tilde{\mathbf{x}}_i, \tilde{y}_i)\}_{i=1}^{N_{\rm tr}}$, a test image $\mathbf{x}$, a pretrained foundation model $\mathcal{F} = f_{L} \circ \ldots \circ f_1$.
    \State{\texttt{// Stage I: search for the optimal intermediate layer}}
    \State Initialize an empty list $\mathcal{P} \leftarrow \{\}$.
    \For{$i = 1$ \textbf{to} $N_{\rm tr}$}
        \For{$\ell$ \textbf{to} $L$}
            \State $\hat{p} \leftarrow S(\tilde{\mathbf{x}}_i, \epsilon(\tilde{\mathbf{x}}_i), l)$ as shown in Equation \ref{eq:metric}
            \State $\mathcal{P} \leftarrow \mathcal{P} \cup \{\hat{p}\}$
        \EndFor
    \EndFor
    \State{$\ell_{\rm opt} \leftarrow \underset{\ell}{\text{argmax}} \, \text{AUROC}(\mathcal{P}, \{\tilde{y}_i\}_{i=1}^{N_{\rm tr}})$}
    \State{\texttt{// Stage II: inference with the optimal intermediate layer}}
    \State{Make a prediction using $\ell_{\rm opt}$ as shown in Equation \ref{eq:infer}} 
    \end{algorithmic}
\end{algorithm}

The defocus blur is employed as the default perturbation, which is a deterministic optical process that redistributes light according to a specific pattern (the blur kernel). It is evident that high-level semantic information and broad scene structures (e.g. the outline of a person or the position of a building) are generally preserved. However, it has been observed that fine-grained details such as textural elements or small textures are frequently attenuated.

\subsection{Revisiting Image Embeddings for AI-generated Image Detection}

Prior training-free methods, such as RIGID \cite{he2024rigid} and MINDER \cite{tsai2024understanding}, rely on the final layer's output embedding and therefore omit the search process. Theoretical analysis \cite{he2024rigid} reveals that the gradient of the cosine similarity is related to the introduced random (Gaussian) perturbation. Foundational image models trained on real images are generally more robust in the embedding space of real images and less robust in AI-generated images. Hence, the difference in the robustness of embedding spaces can be utilized as a metric to detect AI-generated images.

Considering the dimensionality analysis of the intrinsic dimension \cite{recanatesi2019dimensionality,ansuini2019intrinsic}, the output embedding involves both the feature expansion in earlier layers and the feature selection in later layers. Feature selection collapses redundancy and irrelevant variation for downstream tasks. However, the compressed dimension that are redundant and irrelevant to downstream tasks, might be beneficial to detect nuances between real and AI-generated images. Intermediate embeddings can contain richer representation that is beneficial for the detection of AI-generated images.

\begin{figure*}[t]
    \centering
    \includegraphics[width=\linewidth]{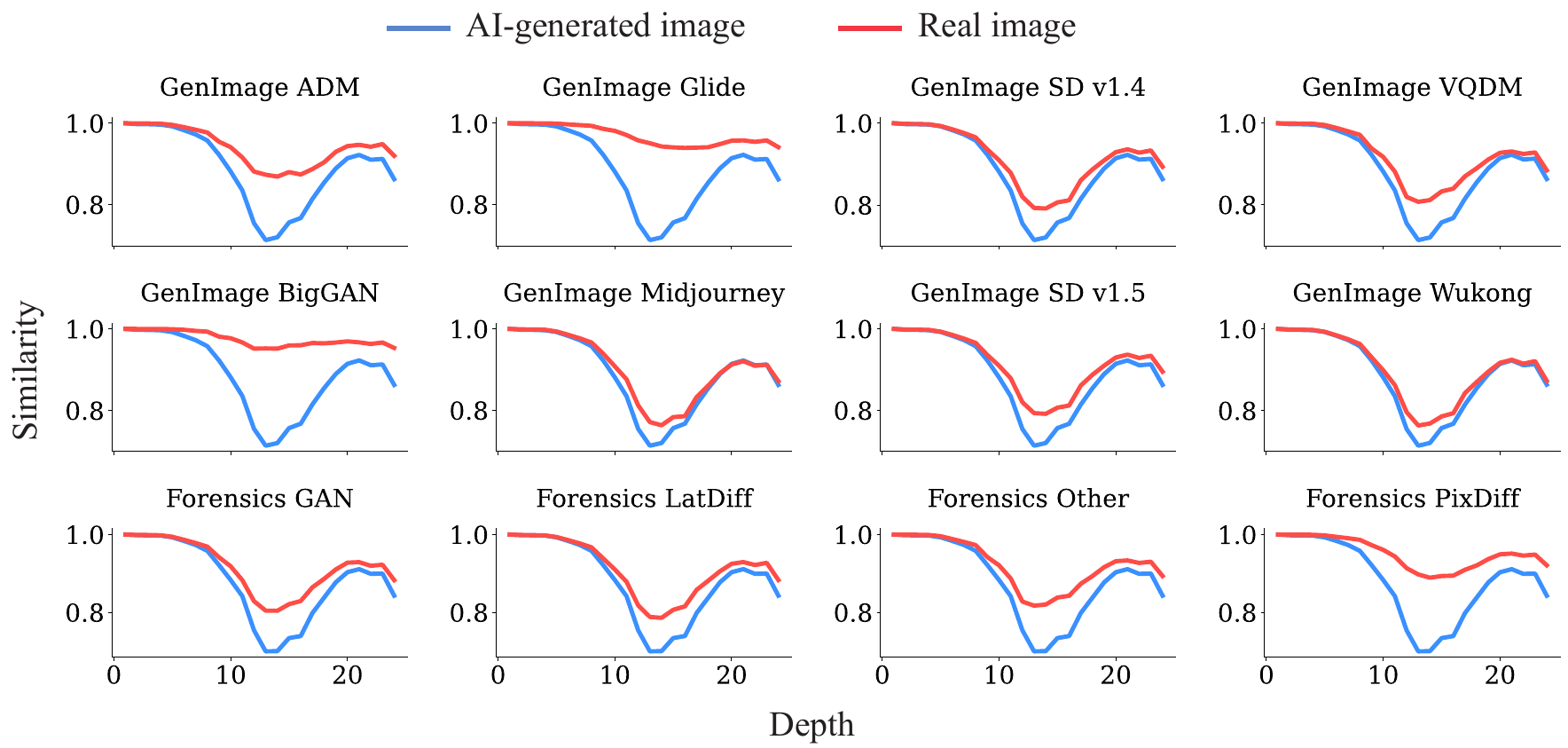}
    \caption{Average cosine similarity profile over model depth for the datasets on the GenImage benchmark and the Forensics Small benchmark. There is an optimal intermediate layer that maximizes the difference in the robustness of image embedding space for real and AI-generated images.}
    \label{fig:sim_profile}
\end{figure*}

Figure~\ref{fig:sim_profile} shows the cosine similarity between pairs of real images and AI-generated images on the GenImage benchmark \cite{zhu2023genimage}  and the Forensics Small benchmark \cite{park2025community}. We compute the average similarity for each layer $\bar{S}(\ell) = \frac{1}{N_{\rm ev}} S(\mathbf{x}_i, \epsilon(\mathbf{x}_i), \ell)$, where $N_{\rm ev} = \lvert \mathcal{D}_{\rm eval} \rvert$. There is a consistent largest similarity difference between real image pairs and AI-generated image pairs in the intermediate layer. The gap in the robustness between real image embeddings and AI-generated image embeddings can be utilized to detect AI-generated images. There is a larger gap in the optimal  intermediate layer compared to the output layer.

\begin{figure}[t]
    \centering
    \includegraphics[width=.92\linewidth]{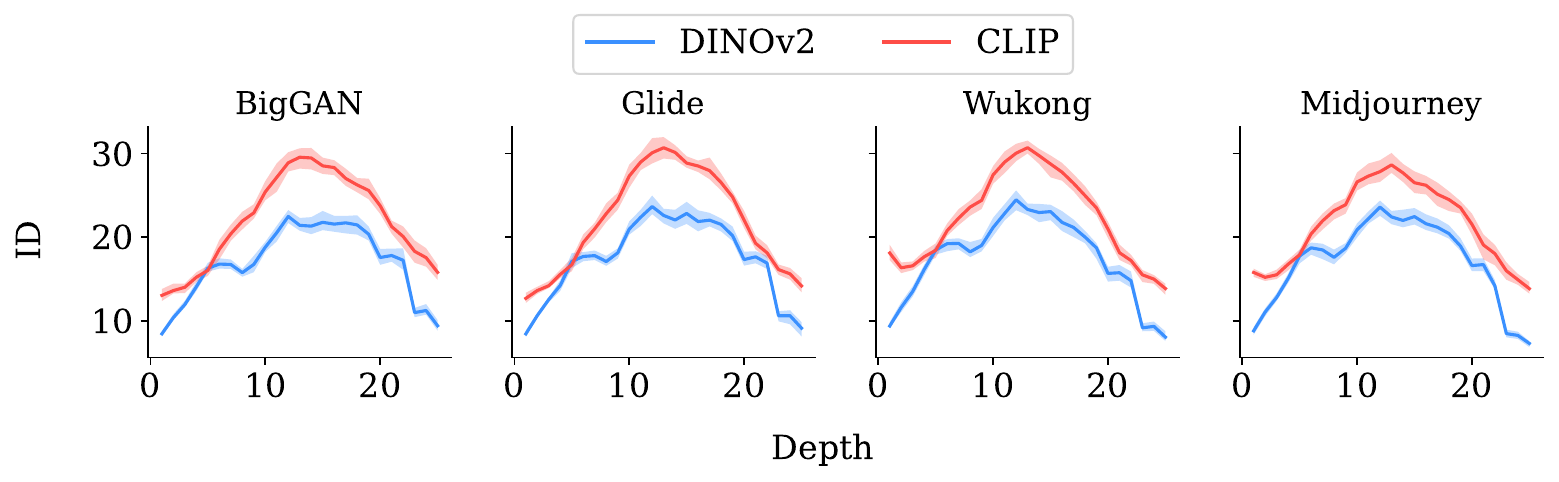}
    \vspace{-1em}
    \caption{Intrinsic Dimension (ID) analysis of data representation manifolds in the image foundation models: DINOv2 \cite{oquab2023dinov2} and CLIP (ViT-L/14) \cite{radford2021learning}. A typical hunchback shape of the profile of the intrinsic dimension is observed, which indicates more diverse features in intermediate layers.}
    \label{fig:intrinsic_dim}
\end{figure}

\subsection{Understanding the Versatility of Intermediate Representations via Intrinsic Dimension}
\label{subsec_ID}

Intrinsic dimension (ID) is a fundamental geometric property of the data representation manifold in an over-parameterized neural network. It represents the minimal number of coordinates to describe data points without significant information loss. In the learning theory, ID plays a vital role in learning function approximations and non-linear decision boundary determination. The number of required data points grows exponentially with the manifold's ID for learning a manifold \cite{narayanan2010sample}. ID is found to be correlated with adversarial training of neural networks \cite{ma2018characterizing,amsaleg2017vulnerability}. A theoretical analysis indicates that an increase in ID effectively reduces the severity level of the perturbation to move a normal example into the adversarial region \cite{amsaleg2017vulnerability}. By employing ID estimator \cite{facco2017estimating,ansuini2019intrinsic}, we examine ID across layers of the feature extractor. ID is calculated based on the ratio between the distances to the second and first nearest neighbor of each data point \cite{facco2017estimating}. Figure \ref{fig:intrinsic_dim} shows the variation of ID for feature extractors used in this study. There is ID expansion in earlier layers and compression in later layers. The hunchback shape of ID as a function of model depth is interpreted as the feature generation in earlier layers \cite{olshausen1997sparse,babadi2014sparseness} and feature selection in later layers \cite{hinton2006reducing,tishby2018information}.

\begin{table*}[tbh]
    \caption{Comparison of AUROC and AP scores on the GenImage benchmark \cite{zhu2023genimage}. We use the optimal layer $\ell_{\rm opt} = 13$ for computing logits in our method.}\label{tab:perf_genimage}
    \includegraphics[width=\linewidth]{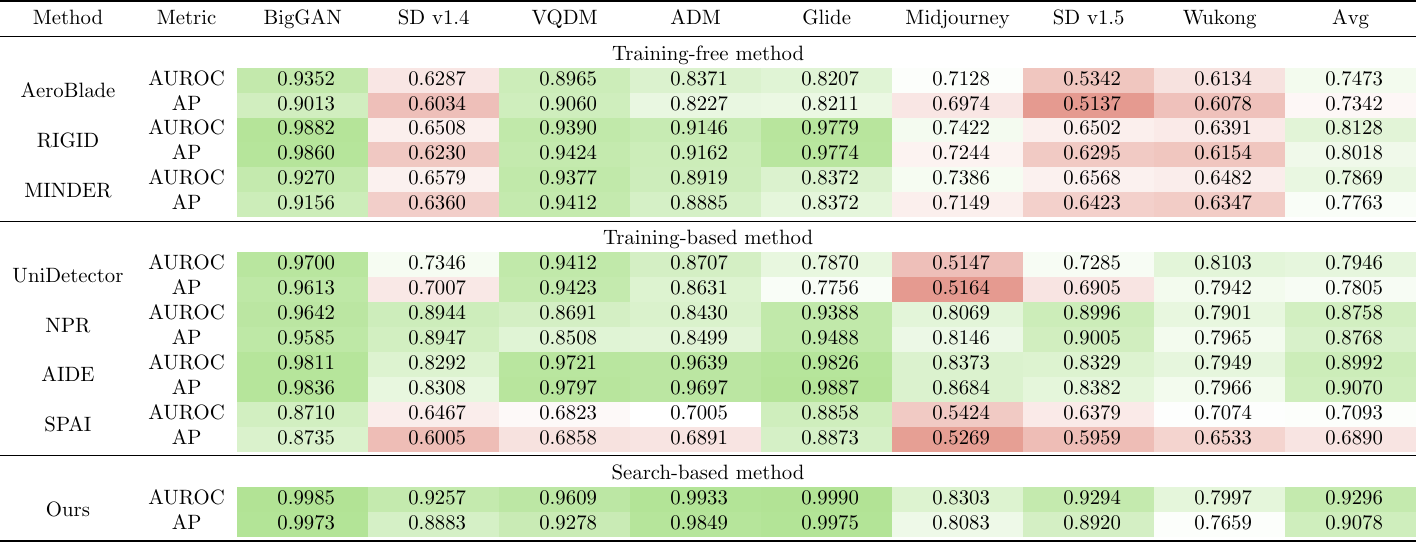}
\end{table*}

\begin{table*}[tbh]
    \caption{Comparison of AUROC and AP scores on the Forensics Small benchmark \cite{park2025community}. We use the optimal layer $\ell_{\rm opt} = 13$ for computing logits in our method.}\label{tab:perf_forensic}
    \includegraphics[width=\linewidth]{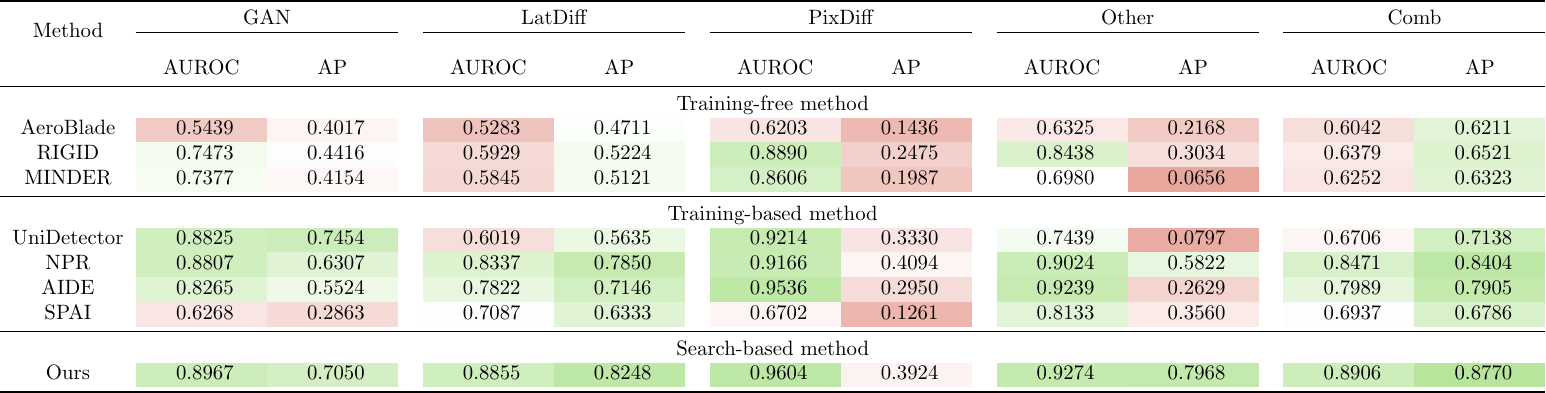}
\end{table*}


The dimensionality analysis indicates that there are more diverse features in intermediate layers than in output layers. Different layers can have different levels of sensitivity to a perturbation. The output layer might not be the most sensitive layer, rendering it sub-optimal in detecting AI-generated images. As shown in Figure \ref{fig:sim_profile}, there is a pronounced variation in cosine similarity across model layers. It indicates that different model layers might have different sensitivity to a perturbation. Besides, the largest difference in cosine similarity between real and AI-generated image embeddings occurs in intermediate layers, indicating that intermediate embeddings might be more expressive compared to the output embeddings that are commonly used for downstream tasks. 

\section{Experiments}

\subsection{Experimental Details}

\paragraph{Datasets} We evaluate the proposed method on two deepfake benchmarks: GenImage \cite{zhu2023genimage} and Forensics Small \cite{park2025community}. The GenImage benchmark consists of a broad range of image classes generated by advanced image generators, including BigGAN \cite{brock2018large}, Stable Diffusion v1.4 and v1.5 \cite{rombach2022high}, VQDM \cite{gu2022vector}, GLIDE \cite{nichol2021glide}, ADM \cite{dhariwal2021diffusion}, Midjourney \cite{midjourney2022} and Wukong \cite{wukong2022}. The Forensics Small benchmark contains $2.78 \times 10^5$ AI-generated images from $4803$ generator models and $2.78 \times 10^5$ real images. Generators are divided into four categories: GAN, LatDiff, PixDiff and Other.

\paragraph{Method Details} We use the CLIP model \cite{radford2021learning} as the feature extractor. We use defocus blur as the perturbation function $\epsilon(\mathbf{x})$. A randomly sampled training dataset from the GenImage benchmark is used to identify the optimal intermediate layer. The size of the randomly sampled subset (from the training set of GenImage) is equal to $30\%$ fraction of the test dataset size. The same intermediate layer is used in the inference process to detect AI-generated images. During the process of identifying the optimal intermediate layer and detecting AI-generated images, both the weight and the architecture are the same as the pre-trained model. In other words, our approach does not make modifications to the feature extractor.

\begin{figure}[tbh]
    \centering
    \includegraphics[width=.82\linewidth]{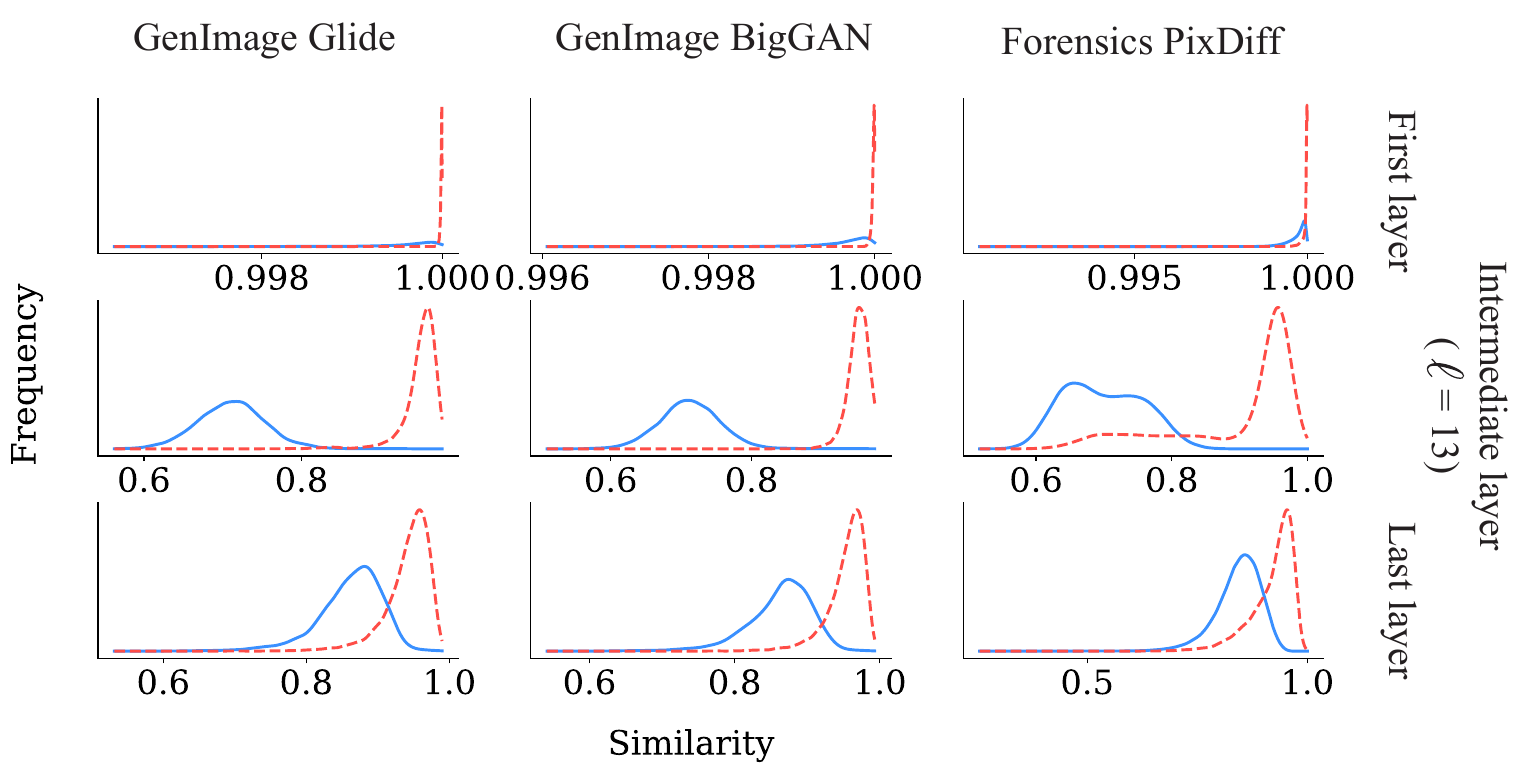}
    \caption{Distribution of cosine similarity between embeddings of AI-generated and real images. Solid curves correspond to AI-generated images while dashed curves correspond to real images. Using intermediate layers improves the separation of embeddings between AI-generated and real images compared to using the first or the last layer.}
    \label{fig:sim_dist}
\end{figure}

\paragraph{Baselines and Metrics} Both training-based and training-free approaches are selected as baselines to examine the proposed method. For training-based methods, UniDetector \cite{ojha2023towards} uses linear probing on the output of the foundational model to detect AI-generated images. NPR \cite{tan2024rethinking}, based on the observation that up-sampling operations produce generalized forgery artifacts, is an artifact representation approach that captures structural artifacts. AIDE \cite{yan2024sanity} utilizes multiple experts to extract visual artifacts and noise patterns for detecting AI-generated images. SPAI \cite{karageorgiou2025any} employs the spectral learning to learn the spectral distribution of real images. Generated images are considered out-of-distribution. For training-free methods, RIGID \cite{he2024rigid} compares the representation similarity between original images and Gaussian noise-perturbed images for detecting AI-generated images. MINDER \cite{tsai2024understanding} improves RIGID by contrastive blurring to increase the distance between perturbed embeddings. Aeroblade \cite{ricker2024aeroblade} considers the difference in the difficulty of reconstructing AI-generated and real images and uses it as the detection metric. We evaluate the performance of AI-generated image detection methods using the AUROC score.

\subsection{Comparison with Baselines}

Table \ref{tab:perf_genimage} and Table \ref{tab:perf_forensic} show the performance comparison for the AI-generated image detection task. Our approach performs favorably against both training-free baselines and training-based baselines. On the GenImage benchmark, our method has the performance gain of 14.37\% on the AUROC score and 13.22\% on the AP score compared to the best training-free baseline, 3.38\% on the AUROC score and same performance (performance gain $<0.10\%$) on the AP score compared to the best training-based baseline. On the Forensics Small benchmark, our method has the performance gain of 39.61\% on the AUROC score and 34.49\% on the AP score compared to best training-free baseline and 5.14\% on the AUROC score and 4.36\% on the AP score compared to the best training-based baseline. The result indicates that our method performs favorably against both the best training-free method and the best training-based method.

\subsection{Intermediate Layer Analysis}

Here, we provide a detailed analysis to study the effect of the intermediate layers on AI-generated image detection.  We extracted embeddings in all layers (\ie $1 \le \ell \le L$). The cosine similarity is computed to predict whether an image is AI-generated as indicated. The AUROC score is used as the metric to examine the prediction performance. In general, the representations
of earlier layers do not provide good separation between real and AI-generated images. While the embedding of the final layer is often used in vision tasks such as image classification, our observation indicates that using an intermediate layer in our method usually achieves the optimal detection performance.

In Figure~\ref{fig:sim_dist}, we visualize the distribution of cosine similarity of the first layer, the optimal intermediate layer, and the last layer. Solid curves correspond to AI-generated images while solid curves correspond to real images. When using the first layer and the last layer, it is difficult to accurately differentiate real and AI-generated images due to the overlap in the distribution. Using intermediate layers, however, improves the separation between distributions of AI-generated and real images. Even though the output layer embeddings are commonly used for downstream tasks such as image classification, they are not optimal for detecting AI-generated images. The ID analysis reveals that when it comes to last few layers, ID decreases and embeddings become less expressive. Nuances between real images and AI-generated images might become undetectable.

\begin{figure*}[tbh]
    \centering
    \includegraphics[width=\linewidth]{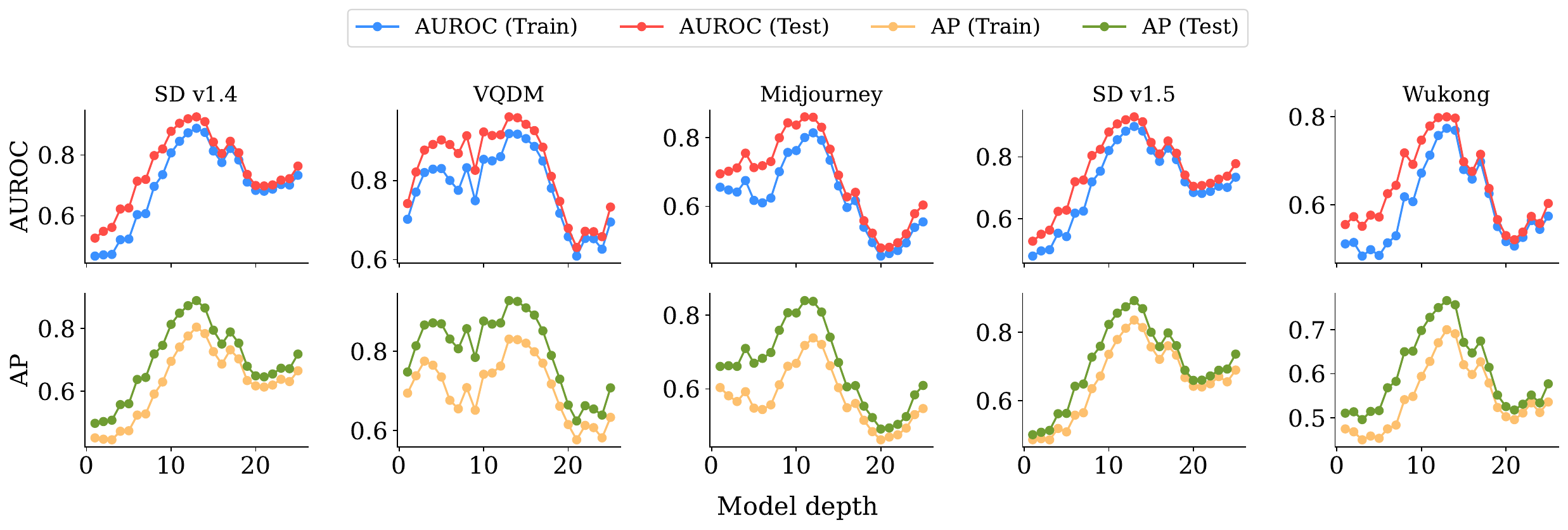}
    \caption{AUROC score and AP score as a function of model depth for images in the training dataset and the test dataset. The training dataset be approximate the test dataset well in terms of identifying the optimal intermediate layer.}
    \label{fig:score_profile}
\end{figure*}

We use the training dataset to determine the optimal intermediate layer. The optimal intermediate layer on the training dataset is used for inference on the test dataset. Figure~\ref{fig:score_profile} shows the discrepancy in the selection of the optimal intermediate layer between the training dataset and the test dataset. There is a significant similarity in the optimal layer selection between these two dataset across different types of AI-generated images and different metrics (including the AUROC score and the AP score), which justifies the practice of utilizing the training dataset to obtain the optimal intermediate layer for the test dataset.

\section{Ablation Study}

\paragraph{Feature extractor} We examine the performance of our proposed method using different image foundation models as feature extractors. Table~\ref{tab:backbone} shows the performance comparison on the GenImage benchmark and the Forensics Small benchmark. When using DINOv2 \cite{oquab2023dinov2} to extract features instead of using the CLIP model \cite{radford2021learning} (ViT-L/14), there is a performance degradation. The improvement of the CLIP model over the DINOv2 model can be attributed to the intrinsic dimension analysis in Section \ref{subsec_ID}, where we show CLIP has a higher intrinsic dimension than DINOv2, offering more versatile intermediate representations for AI-generated image detection.

\begin{table}[t]
    \centering
    \caption{Performance comparison of applying different types of perturbations. The average performance is reported on two benchamrks: the GenImage benchmark and the Forensics Small benchmark.}
    \begin{tabular}{cccccc}
        \toprule
        \multirow{2}{*}{Perturbation} & \multicolumn{2}{c}{GenImage} &  & \multicolumn{2}{c}{Forensics Small} \\
        \cline{2-3}\cline{5-6}
              & AUROC & AP & & AUROC & AP\\
        \midrule
        Contrast          & 0.8439 & 0.8266 & & 0.8311 & 0.8215 \\
        Elastic transform & 0.8683 & 0.8599 & & 0.8854 & 0.8868 \\
        JPEG compression  & 0.8196 & 0.8008 & & 0.7785 & 0.7916 \\
        Impulse noise     & 0.7648 & 0.7553 & & 0.6105 & 0.6398 \\
        Gaussian noise    & 0.6737 & 0.6402 & & 0.7184 & 0.7320\\
        Defocus blur      & 0.9296 & 0.9078 & & 0.8906 & 0.8770 \\
        Shot noise        & 0.7320 & 0.7047 & & 0.7079 & 0.7314 \\
        Zoom blur         & 0.9090 & 0.8974 & & 0.8633 & 0.8431 \\
        \bottomrule
    \end{tabular}
    \label{tab:ablat_perturb}
\end{table}

\begin{table*}[tbh]
    \caption{Ablation study on the effect of the feature extractor on the detection performance on the GenImage benchmark and Forensics Small benchmark. Two feature extractors are examined: CLIP and DINOv2. Both feature extractor use ViT-L/14 as the backbone model.}\label{tab:backbone}
    \includegraphics[width=\linewidth]{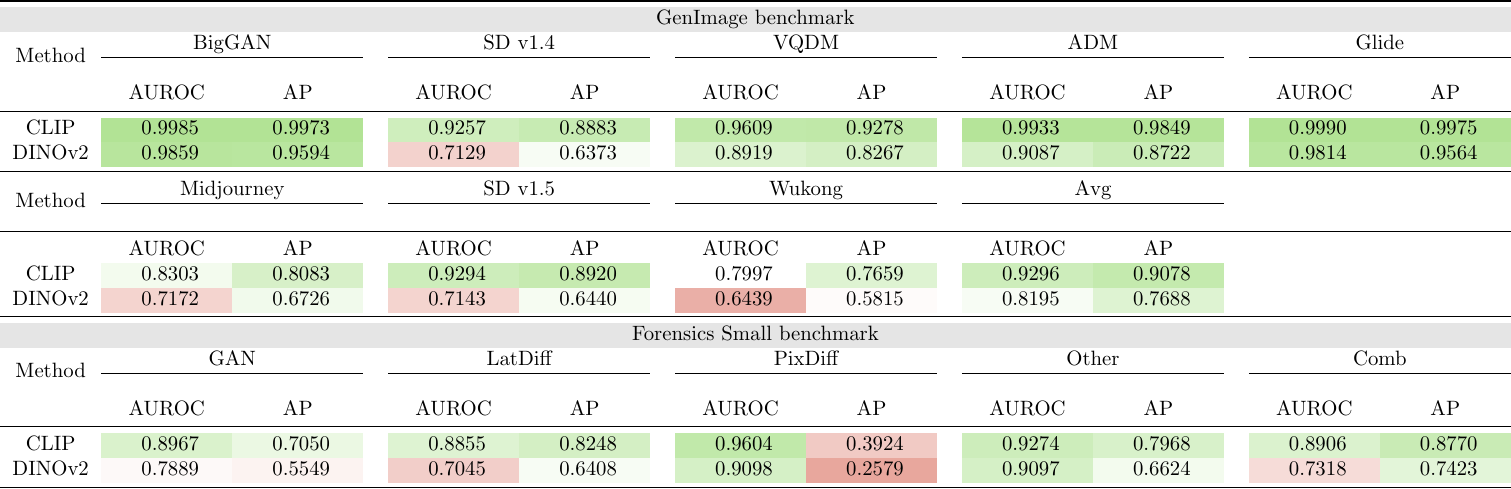}
\end{table*}

\begin{figure}[tbh]
    \centering
    \includegraphics[width=.88\linewidth]{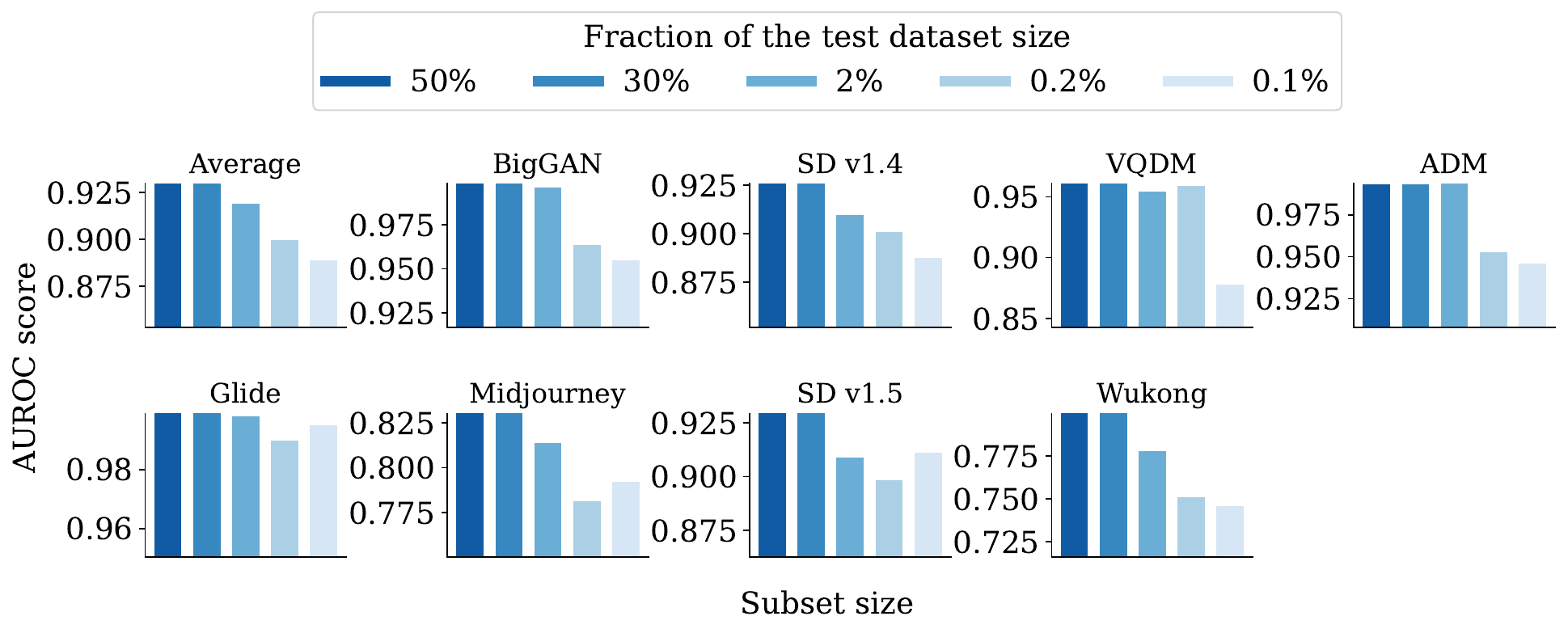}
    \caption{Effect of the size of the randomly sampled training dataset on the detection performance for AI-generated images. We randomly sample the number of training images that is equal to $k\%$ of the test dataset size. $k \in \{50, 30, 2, 0.2, 0.1\}$.}
    \label{fig:ablation_fraction}
\end{figure}

\paragraph{Size of search dataset} We use the training dataset to search for the optimal intermediate layer. We empirically find that a randomly sampled subset of the training dataset is sufficient to identify the optimal layer. We examine the effect of the size of the randomly sampled subset, and Figure~\ref{fig:ablation_fraction}  shows the effect of the subset size. In our pipeline, the size of the randomly sampled subset of the training dataset is equivalent to $30\%$ of the test dataset size. When further increasing the size of the randomly sampled subset, the same optimal layer is determined. When decreasing the size of the randomly sampled subset, on the other hand, the identified layer can be sub-optimal.

\paragraph{Perturbation type} We examine the effect of different types of perturbations on the detection performance. 8 perturbation types are examined: Contrast, Elastic transformation, JPEG compression, Impulse noise, Gaussian noise, Defocus blur, Shot noise and Zoom blur. For a detailed description of 8 perturbation methods, we refer the reader to \cite{hendrycks2019benchmarking}. The exaggerated visualization of different types of perturbations is shown in Appendix \ref{append:perturb}. Average AUROC scores and AP scores are reported in Table~\ref{tab:ablat_perturb}. Overall, applying defocus blur perturbation results in the best performance. This finding can be attributed to the fact that defocus blur preserves the global structure and tends to attenuate fine-grained details, thereby encouraging the detector to focus on structural differences.

\section{Conclusion}

In this paper, we propose a novel search-based approach for detecting AI-generated images. By searching for the optimal intermediate layer to obtain the most separable similarity features between the original images and perturbed images, our approach improves the detection performance over state-of-the-art training-based and training-free methods.
We provide comprehensive analysis and intrinsic dimension evaluation to explain how the versatility of the intermediate representations derived from a pretrained image foundation model can be used to design powerful AI-generated image detectors. Our results reveal that the commonly used output embeddings often lead to sub-optimal performance in detecting AI-generated images using an image foundation model.
Our method can be used with any off-the-shelf image foundation model to extract intermediate representations. Hence, we believe the detection performance can scale with the representation learning capability of future image foundation models.

\paragraph{Impact Statement} 
This work focuses on developing a reliable method to address the problem of detecting AI-generated images, with the aim of mitigating risks posed by generative models including copyright violation, AI autophagy, etc. The work can be applied to enhance the reliability of media forensics and support trustworthy information dissemination. The proposed approach does not involve the generation of harmful or offensive content.

\paragraph{Limitations.}

The search process in our proposed method does not involve any model fine-tuning. The performance might be improved by using search and fine-tuning together. We choose to keep the model parameter frozen in this work, in order to show the competitive performance of the search-based method compared to state-of-the-art methods.

\bibliography{refs}
\bibliographystyle{plain}

\newpage
\appendix
\onecolumn

\section{Implementation Details}
\label{append:implement_detail}

We use pretrained CLIP to extract features. Besides, we test the performance of using DINOv2 as the feature extractor. Both models use ViT-L/14 as the backbone model. Images are resized to $224 \times 224$ and then used as the input to the foundational vision model. We use the training dataset on the GenImage benchmark to identify the optimal layer.

\subsection{Baseline Implementation}

In RIGID and MINDER baselines, the DINOv2 model is used to detect AI-generated images. We use model weights fine-tuned on the GenImage benchmark in the model inference process for NPR and AIDE baselines. UniDetector trains a classification layer using the curated dataset \cite{wang2020cnn}. The model weight of SPAI is obtained by training on the curated dataset, where AI-generated images are generated by latent diffusion model \cite{rombach2022high} while real images are collected from the publicly available dataset \cite{corvi2023detection}.

\section{Perturbations}
\label{append:perturb}

\begin{figure}[htb]
    \centering
    \includegraphics[width=.7\linewidth]{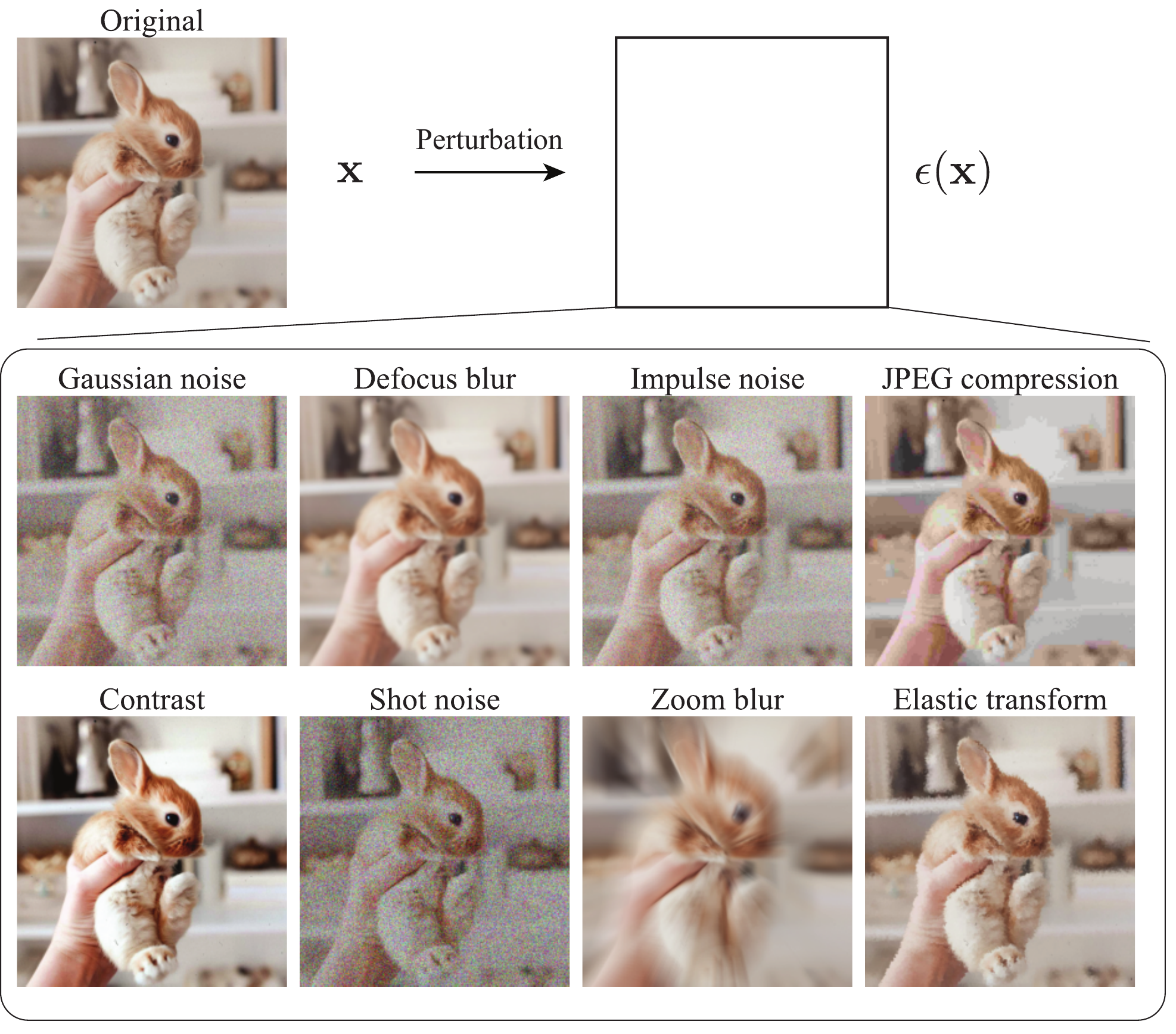}
    \caption{Algorithmically generated corruptions to apply perturbation to input images. Each perturbation type has eight severity levels. We use the cosine similarity between the embeddings of the original image and the perturbed image to make a binary classification on whether the original image is an AI-generated image (\ie, AI-generated image). Perturbations are exaggerated for better visualization purposes.}
    \label{fig:perturb}
\end{figure}

Following \cite{hendrycks2019benchmarking}, we apply different perturbation types with different severity levels to input images. Figure \ref{fig:perturb} shows eight different perturbation types: Gaussian noise, defocus blur, impulse noise, JPEG compression, contrast, shot noise, zoom blur and elastic transform. We use exaggerated severity levels to visualize the effect of different perturbations on original images.

\begin{figure}[tbh]
    \centering
    \includegraphics[width=.88\linewidth]{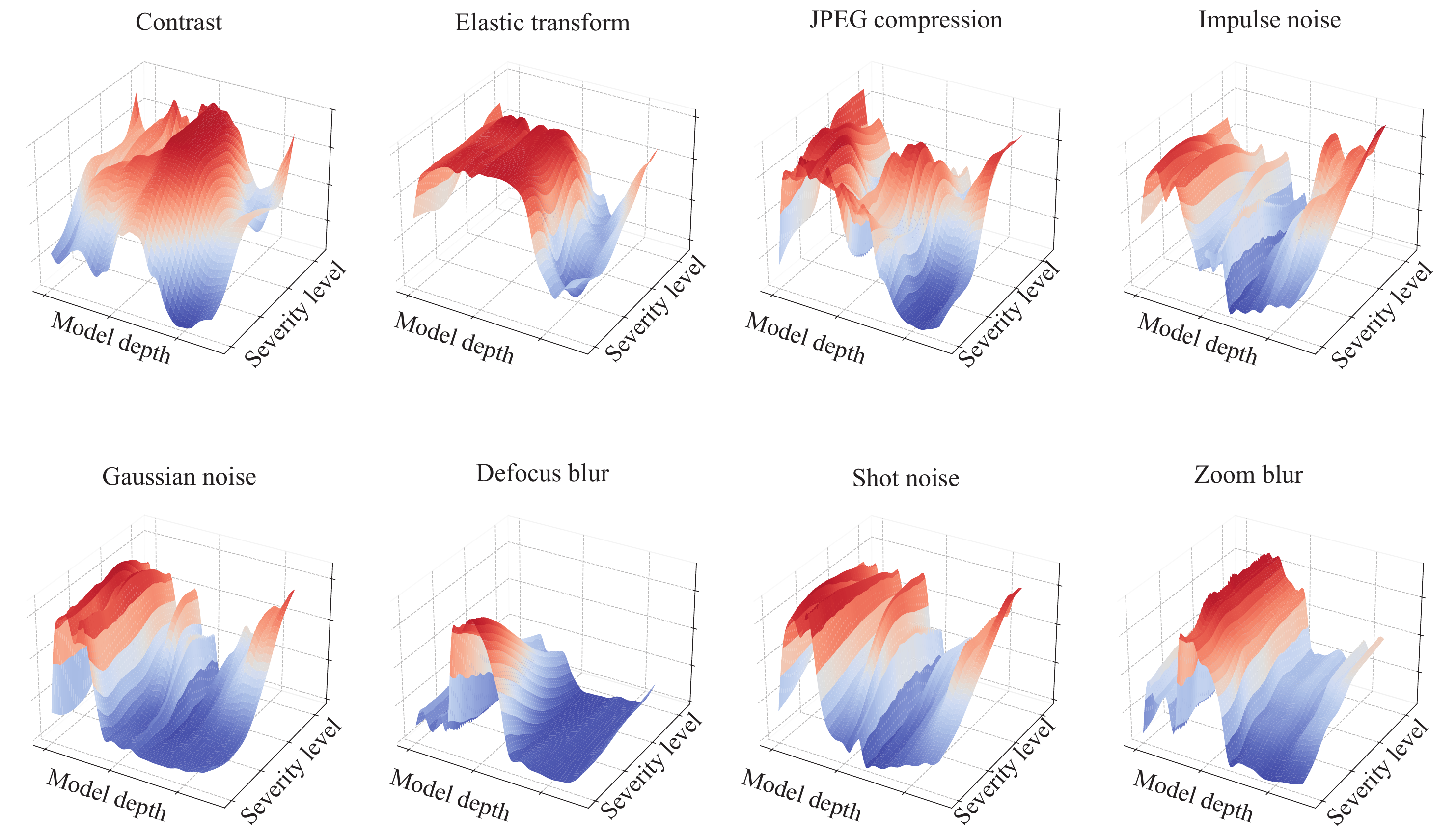}
    \caption{Variation of AUROC score (Z axis) as a function of model depth and severity level for different types of perturbations on the GenImage benchmark.}
    \label{fig:perturb_surface}
\end{figure}

Figure~\ref{fig:perturb_surface} shows the effect of model depth and severity on the detection performance. For each dataset, we examine the same fixed search space (model dpeth, perturbation type and severity level). We empirically find that using the perturbation of defocus blur with a severity level of 7 to apply the perturbation function $\epsilon(\mathbf{x})$ leads to the best detection performance. Besides, the optimal intermediate layer can very when different types of perturbations are applied. We identify the optimal intermediate layer when different types of perturbations are applied. The result is shown in Table~\ref{tab:opt_layer}.

\begin{table}[tbh]
    \centering
    \caption{Optimal intermediate layer for different types of perturbations. The training dataset is used to determine the optimal intermediate layer.}
    \label{tab:opt_layer}
    \resizebox{\linewidth}{!}{%
    \begin{tabular}{ccccccccc}
        \toprule
        Perturbation type          & Contrast & Elastic transform & JPEG compression & Impulse noise \\
        \midrule
        Optimal intermediate layer & 13 & 13 & 24 & 24 \\
        \midrule
        Perturbation type          & Gaussian noise & Defocus blur & Shot noise & Zoom blur \\
        \midrule
        Optimal intermediate layer & 24 & 13 & 24 & 13 \\
        \bottomrule
    \end{tabular}}
\end{table}

\section{Representation Across Model Depth}

\subsection{Detection performance As a Function of Model Depth}

\begin{figure}[tbh]
    \centering
    \includegraphics[width=.88\linewidth]{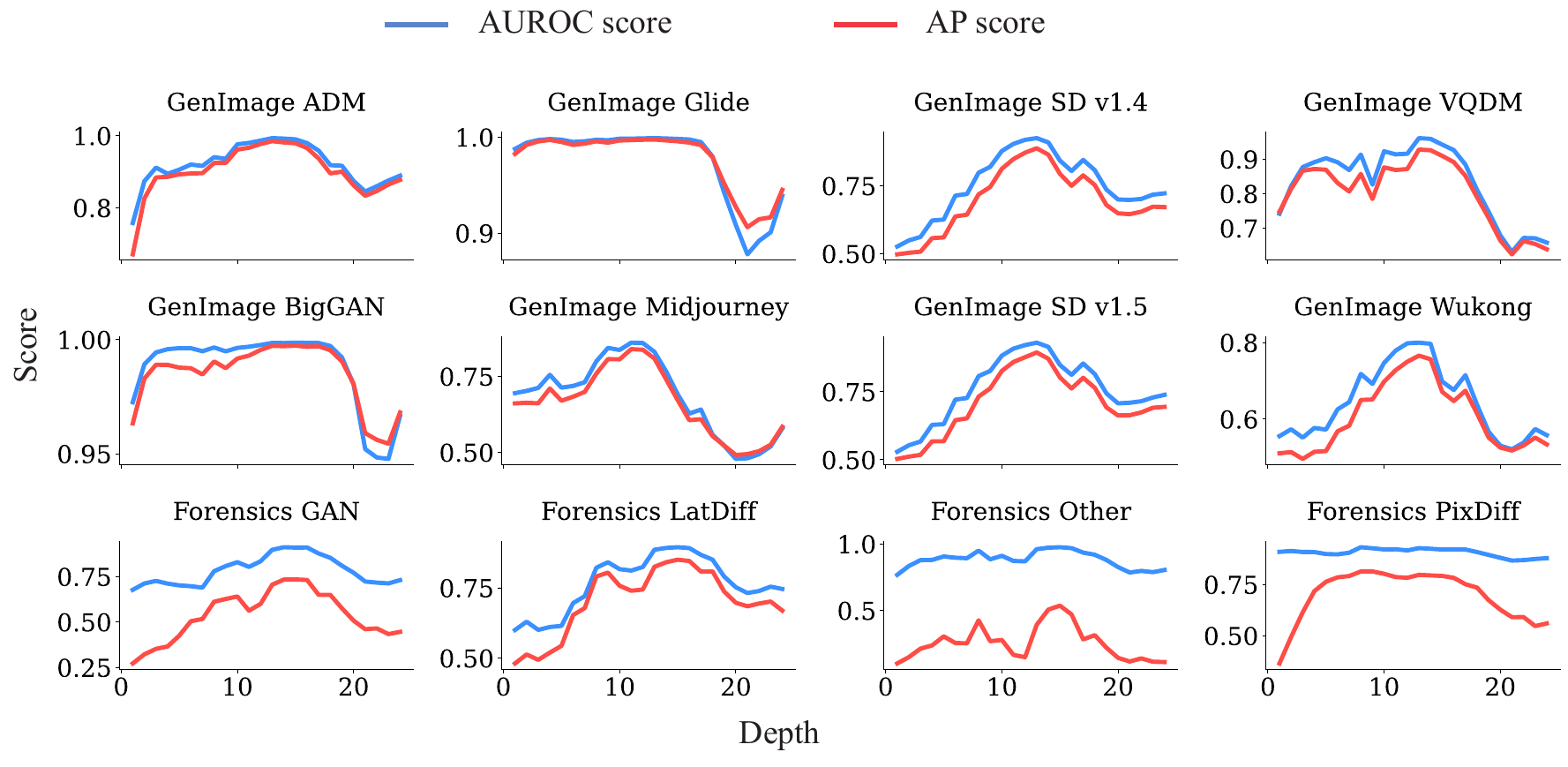}
    \caption{AUROC score and AP score on the GenImage benchmark and Forensics Small benchmark as a function of model depth. Intermediate layers can exhibit better detection performance compared to the output layer.}
    \label{fig:auroc_profile}
\end{figure}

The intrinsic dimension analysis in Figure~\ref{fig:intrinsic_dim} reveals that intermediate layers have more diverse feature representations. The average cosine similarity profile in Figure~\ref{fig:sim_profile} indicates that the difference in the robustness of embedding spaces between real and AI-generated images become largest in the intermediate layer. We examine how does these two features affect the detection performance by computing the AUROC score and the AP score across model depth. Figure~\ref{fig:auroc_profile} shows the profile of the AUROC score and the AP score on the GenImage benchmark and the Forensics Small benchmark. Although output embeddings are typically used for downstream tasks such as image classification, using them to compute prediction logits can be sub-optimal. Besides, the optimal intermediate intermediate layer can vary when using different types of perturbations. 

\section{AI-Generated Image Datasets}

\begin{figure}[tbh]
    \centering
    \includegraphics[width=0.8\linewidth]{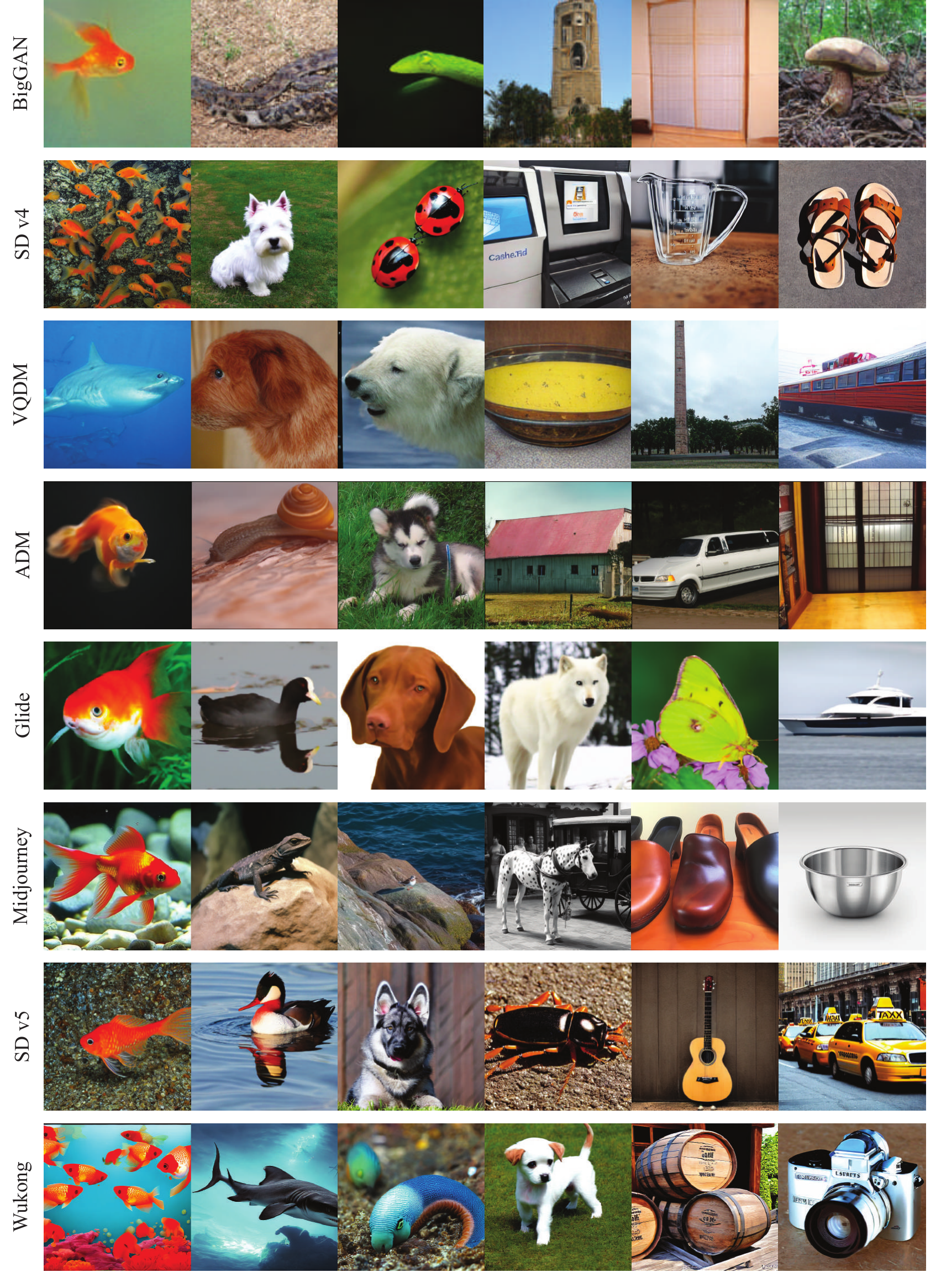}
    \caption{Display of AI-generated images in the GenImage benchmark. Generation models include BigGAN, Stable Diffusion v1.4, VQDM, ADM, GLIDE, Midjourney and Wukong.}
    \label{fig:dataset_genimage}
\end{figure}

Figure~\ref{fig:dataset_genimage} shows examples of AI-generated images in the GenImage benchmark. The GenImage benchmark collects more than one million pairs of AI-generated images and retrieved real images. Advanced diffusion models and GAN models are used to produce AI-generated images. 1000 image labels in the ImageNet dataset \cite{deng2009imagenet} are leveraged to produce AI-generated images.

Table~\ref{tab:genimage_stat} counts the number of real images and AI-generated images on the GenImage benchmark. Binary classes are well balanced on this benchmark.
\begin{table}[tbh]
    \centering
    \caption{Count of the number of images on the GenImage benchmark.}
    \resizebox{\linewidth}{!}{%
    \begin{tabular}{ccccccccc}
        \toprule
        Category            & BigGAN & SD v1.4 & VQDM & ADM & Glide & Midjourney & SD v1.5 & Wukong \\
        \midrule
        Real images         &  6000 & 6000 & 6000 & 6000 & 6000 & 6000 & 8000 & 6000 \\
        AI-generated images &  6000 & 6000 & 6000 & 6000 & 6000 & 6000 & 8000 & 6000 \\
        \bottomrule
    \end{tabular}}
    \label{tab:genimage_stat}
\end{table}

\begin{figure}[tbh]
    \centering
    \includegraphics[width=0.8\linewidth]{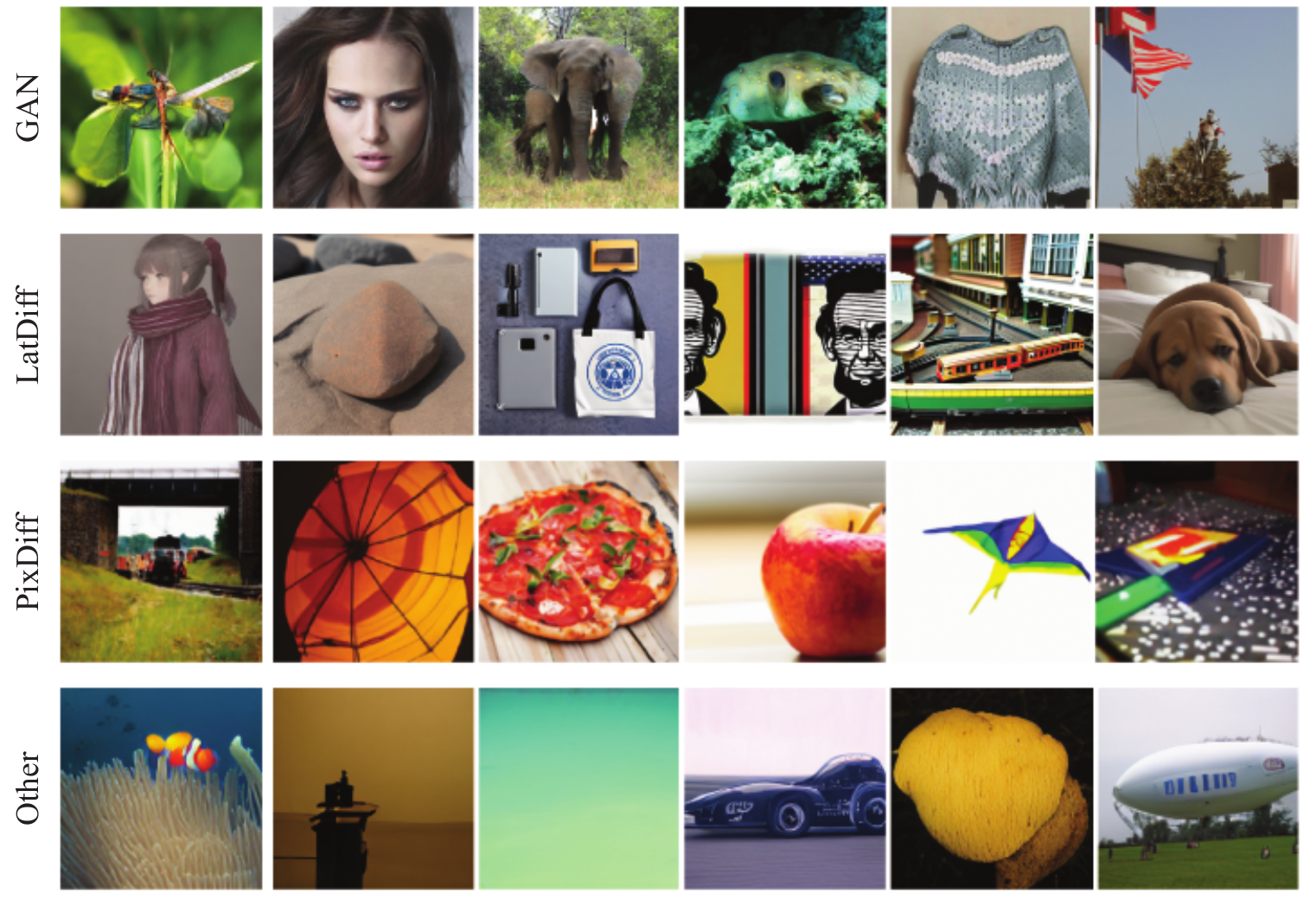}
    \caption{Display of AI-generated images in the Forensic Small benchmark. There are four types of image generators: GAN, LatDiff, PixDiff and other.}
    \label{fig:dataset_forensic}
\end{figure}

Figure~\ref{fig:dataset_forensic} shows examples of AI-generated images in the Forensic Small benchmark. There are $2.78 \times 10^5$ AI-generated images and $2.78 \times 10^5$ real images. Generated images are classified into four categories: GAN, LatDiff, PixDiff and Other. The performance on these four categories is reported in Table \ref{tab:perf_forensic}. To ensure a balanced dataset, we randomly sample real images. The subset of sampled real images has the same size as that of AI-generated images. Comb combines all AI-generated images and all real images.

Table~\ref{tab:forensics_stat} shows the count of the number of real images and AI-generated images. Except for the LatDiff category, there is a strong class imbalance: the number of real images is much larger than that of AI-generated images.
\begin{table}[tbh]
    \centering
    \caption{Count of the number of images on the Forensics Small benchmark.}
    \begin{tabular}{ccccc}
        \toprule
        Category            & GAN & LatDiff & Other & PixDiff  \\
        \midrule
        Real images         & 278096 & 278096 & 278096 & 278096 \\
        AI-generated images & 57397  & 207581 & 6216   & 7251   \\
        \bottomrule
    \end{tabular}
    \label{tab:forensics_stat}
\end{table}

\end{document}